\documentclass{ieeeaccess}
\usepackage{cite}
\usepackage{amsmath,amssymb,amsfonts}
\usepackage{algorithmic}
\usepackage{graphicx}
\usepackage{textcomp}
\usepackage{listings}
\usepackage{url}
\usepackage{hyperref}
\usepackage{multirow}
\usepackage{hhline}
\usepackage{enumitem}
\usepackage{stfloats}
\usepackage{etoolbox}
\usepackage{soul}
\usepackage{caption}

\usepackage[para,online,flushleft]{threeparttable}
\def\BibTeX{{\rm B\kern-.05em{\sc i\kern-.025em b}\kern-.08em
    T\kern-.1667em\lower.7ex\hbox{E}\kern-.125emX}}
\begin{document}
\history{}
\doi{10.1109/ACCESS.2024.3358206}

\title{Socially Aware Synthetic Data Generation for Suicidal Ideation Detection Using Large Language Models}
\author{\uppercase{Hamideh Ghanadian}\authorrefmark{1}, 
\uppercase{Isar Nejadgholi}\authorrefmark{2}, 
\uppercase{Hussein Al Osman}\authorrefmark{3}}
\address[1]{University of Ottawa, Ottawa, Canada (e-mail: Hghan053@uottawa.ca)}
\address[2]{National Research Council Canada, Ottawa, Canada  (e-mail: Isar.nejadgholi@nrc-cnrc.gc.ca)}
\address[3]{University of Ottawa, Ottawa, Canada (e-mail: Hussein.alosman@uottawa.ca)}

\markboth
{Ghanadian \headeretal: Preparation of Papers for IEEE TRANSACTIONS and JOURNALS}
{Ghanadian \headeretal: Preparation of Papers for IEEE TRANSACTIONS and JOURNALS}

\corresp{Corresponding author: Hamideh Ghanadian (e-mail: Hghan053@ uOttawa.ca)}

\begin{abstract}
Suicidal ideation detection is a vital research area that holds great potential for improving mental health support systems. However, the sensitivity surrounding suicide-related data poses challenges in accessing large-scale, annotated datasets necessary for training effective machine learning models. To address this limitation, we introduce an innovative strategy that leverages the capabilities of generative AI models, such as ChatGPT, Flan-T5, and Llama, to create synthetic data for suicidal ideation detection. Our data generation approach is grounded in social factors extracted from psychology literature and aims to ensure coverage of essential information related to suicidal ideation.
In our study, we benchmarked against state-of-the-art NLP classification models, specifically, those centered around the BERT family structures. When trained on the real-world dataset, UMD, these conventional models tend to yield F1-scores ranging from 0.75  to 0.87. Our synthetic data-driven method, informed by social factors, offers consistent F1-scores of 0.82 for both models, suggesting that the richness of topics in synthetic data can bridge the performance gap across different model complexities. Most impressively, when we combined a mere 30\% of the UMD dataset with our synthetic data, we witnessed a substantial increase in performance, achieving an F1-score of 0.88 on the UMD test set. Such results underscore the cost-effectiveness and potential of our approach in confronting major challenges in the field, such as data scarcity and the quest for diversity in data representation.

\end{abstract}


\titlepgskip=-15pt

\maketitle

\section{Introduction}\label{sec:introduction}
\PARstart{A}{ccording} to the World Health Organization\footnote{\href{https://www.who.int/news-room/fact-sheets/detail/suicide}{The World Health Organization (WHO)}} more than 700,000 people die due to suicide every year. Suicide remains a global health crisis, accounting for a significant proportion of mortality rates across various age groups. Suicidal ideation, often a precursor to actual suicide attempts, involves the presence of persistent thoughts, contemplation, or planning related to self-harm or death. Early identification of suicide ideation and intervention to protect individuals at risk of suicide are crucial steps in reducing suicide rates and providing appropriate mental health support. Early detection of suicidal ideation is a complex task, as it requires the integration of various factors, including psychological, social, and environmental variables\cite{de2018understanding}.

In recent years, the proliferation of digital platforms and social media has provided an unprecedented opportunity to capture and analyze large-scale data related to mental health \cite{kumar2023predicting} \cite{raza2022predicting}. Machine learning and Natural Language Processing (NLP) techniques have shown promise in detecting linguistic patterns and indicators of suicidal ideation in diverse text-based data sources, such as social media posts, online forums, and electronic health records\cite{abdulsalam2022suicidal},\cite{li2022deep},\cite{kodati2023identifying},\cite{sadiq2022predicting}.

However, the use of machine learning technologies requires high volumes of data. Data collection and annotation processes are time-consuming and impose significant financial costs \cite{wei2018clinical}. Specifically, obtaining a substantial amount of labeled data related to suicide can be challenging and limited due to several factors inherent to the nature of suicide research. The sensitive and stigmatized nature of suicide often presents barriers to data collection. Individuals and organizations may be rightfully reluctant to share personal or confidential information related to suicide, fearing potential negative consequences. 

Synthetic data generation offers a viable solution to mitigate the data availability limitation by creating artificially generated data that closely resembles real-world data. Synthetic data generation can be instrumental in machine learning applications as it addresses many challenges of real data collection and annotation. Here we review a list of common challenges in data collection that can be managed through synthetic data generation.  

\vspace{5pt}
\noindent\textbf{Data Scarcity:} In many NLP tasks, such as mental health-related applications, there may be limited availability of relevant data due to privacy concerns or the complexity and cost associated with manual annotation. Synthetic data generation allows researchers and practitioners to overcome data scarcity issues and augment the limited amount of publically available data \cite{babbar2019data}.

\vspace{5pt}
\noindent\textbf{Data Diversity:} NLP models trained on limited data may suffer from poor generalization and performance when exposed to diverse and previously unseen examples. Moreover, in real data, certain topics can be undermined or overlooked due to being less discussed. This can happen for several reasons. For example, certain topics may be stigmatized and considered too sensitive or taboo, making people hesitant to openly discuss them. This could include subjects related to mental health, addiction, discrimination, or social issues that carry societal stigmas. Additionally, topics relevant to marginalized or minority communities may receive less discussion due to systemic biases, unequal representation, or limited platforms for their voices to be heard. Also, some topics may be highly specialized or complex, requiring specific expertise or background knowledge to engage in meaningful discussions.
Encouraging diverse perspectives and actively seeking out less-discussed topics can contribute to a more comprehensive and nuanced understanding of real-world issues. Synthetic data generation can help enrich the training data by introducing a wider variety of linguistic patterns, sentence structures, vocabulary and topics. This, in turn, improves the model's ability to handle variations in natural language and increases its robustness \cite{nikolenko2021synthetic}.

\vspace{5pt}
\noindent\textbf{Privacy Preservation:} Suicide detection tasks often involve sensitive information. Generating synthetic data allows researchers to create representative samples that preserve the privacy of individuals while maintaining the statistical properties and distribution of the original data. \cite{lu2023machine}

\vspace{5pt}
\noindent\textbf{Annotation Cost:} Suicide detection is a complex task, and high-quality annotation can only be performed by experts and trained annotators, which can be costly \cite{chau2020understanding}. Synthetic data generation addresses the data annotation issue by targeted data generation so that each generated example is pre-labeld with a specific category. Although these labels might be noisy to some extent, they might be preferable in some settings as they come at no additional cost.\medskip

To investigate the feasibility and effectiveness of synthetic data generation in the task of suicide ideation generation, we use Generative Large Language Models (GLLMs) for data synthesis and use the generated data to train/test text classifiers. To train classifiers, we fine-tune pretrained BERT-like Large Language Models (LLMs) as state-of-the-art text classifiers.

To enhance the quality of the generated data, we benefit from domain knowledge from psychology. Previous research highlights the importance of incorporating social factors in the design process of NLP systems \cite{hovy-yang-2021-importance}. Specifically, when generating data with LLMs, external sources of domain knowledge can be leveraged to guide the data generation process \cite{hu-etal-2022-empowering}. For the task of suicidal ideation detection, such knowledge can be drawn from a vast body of research in psychology devoted to gaining an understanding of the social factors associated with suicidal ideation and behavior.
In this work, we review the psychology literature to extract the social factors tightly tied to suicidal ideation and 
use this knowledge for more effective prompt engineering when generating data with GLLMs. Guiding the data generation with these factors enables the creation of diverse and representative examples of suicidal ideation.\medskip

\noindent The main contributions of this study are as follows:

\begin{itemize}[leftmargin=*]
\item We extracted the relevant social factors associated with suicidal ideation through a comprehensive review of existing literature, research papers and clinical studies to identify key themes related to suicidal ideation. These themes encompass a wide range of factors, including risk factors, common triggers and mental health indicators. Leveraging a socially aware data synthesis approach, we pave the way for more accurate and reliable suicidal ideation detection systems.

\item Our study examines three GLLMs' performance in producing synthetic datasets with Zero-Shot and Few-Shot learning techniques. Utilizing the ChatGPT, Flan-T5, and Llama 2 models and leveraging the extracted social factors from the psychology literature, we generated nine datasets with diverse characteristics. 

\item We trained classifiers by fine-tuning two pre-trained language models, ALBERT and DistilBERT, using the generated datasets. We tested these models on two test sets, the University of Maryland Suicidality dataset (UMD) and a human-annotated synthetic dataset presented in this paper. Our findings indicate that the GLLMs have significant potential for generating a suicide-related dataset comparable with real available datasets such as UMD. More significantly, the integration of social knowledge may significantly enhance the quality of the generated datasets and lead to more robust classifiers.

\item We augmented our best-performing synthetic dataset using subsets of the UMD dataset to evaluate the efficacy of data augmentation in suicidal detection applications. 
Our results show that models trained with synthetic data augmented with a small set of real-world data can outperform models trained by large annotated real-world datasets.
\end{itemize}

This paper is organized as follows: In Section \ref{sec:background}, we review the literature and related background. In Section \ref{sec:methodology}, we explain our methodology for generating and evaluating the proposed synthetic data generation, specifics of the classifiers and datasets we use in this work. Section \ref{sec:Results} presents our results, and Section \ref{sec:Discussion} discusses these results in detail. Additionally, the conclusion and  possible future works are discussed in Section \ref{sec:future-works}. We complete the article by including an ethical statement in Section \ref{sec:Ethics}, which delves into the ethical aspects and considerations associated with our work. 

\section{Background and Related Work}\label{sec:background}
In this section, we review the related work in suicidal ideation detection in psychology as well as NLP research that addresses the task of suicide detection. We also review the previous works that focused on generating synthetic data for a variety of NLP tasks.

\subsection{ Suicidal Ideation and Related Social Factors }
Suicidal ideation has been a subject of extensive research within the field of psychology. Understanding the underlying elements and risk factors related to suicidal thoughts and behaviors is crucial for developing effective prevention and intervention strategies.

One important area of investigation is the identification of risk factors associated with suicidal ideation. Numerous studies have examined the impact of psychological factors such as depression, anxiety, hopelessness, and feelings of worthlessness on the development of suicidal thoughts\cite{franklin2017risk,bentley2016anxiety}. 
These Studies investigate the strong association between suicidal thoughts and conditions like depression \cite{orsolini2020understanding,kalin2020insights}, bipolar disorder \cite{da2015risk}, borderline personality disorder \cite{paris2019suicidality}, and substance abuse \cite{lee2017mental}. By examining the interplay between these conditions, researchers aim to develop targeted interventions to address the unique challenges faced by individuals struggling with suicidal ideation \cite{okolie2017systematic,kleiman2017examination}.
Additionally, environmental factors such as a history of trauma, social isolation, and access to lethal means have been identified as potential risk factors \cite{leigh2017overview,holt2015loneliness,allchin2019limiting}.

Psychology offers valuable insights into the diverse processes and factors that contribute to suicide risk. Psychological theories and frameworks such as the interpersonal theory of suicide \cite{van2010interpersonal}, the cognitive model of suicidal behavior\cite{wenzel2008cognitive}, and the social-ecological model\cite{cramer2017social} provide a theoretical foundation for understanding the complex interplay between individual vulnerabilities and environmental factors.

The extensive research conducted on suicidal ideation and associated topics in psychology has significantly contributed to the understanding of the complex factors involved. By unraveling the causes, risk factors, and protective factors associated with suicidal thoughts, researchers aim to develop effective prevention strategies, enhance mental health interventions, and ultimately reduce the global burden of suicide. In Section \ref{sec:topics}, we enumerate the social factors that are discussed in the literature as relevant topics to suicidal ideation. 

\subsection{ Suicidal Ideation Detection Using NLP}
In recent years, there has been a growing interest in using NLP techniques for suicide prevention \cite{fernandes2018identifying, bejan2022improving}.
Researchers have developed suicide detection systems to analyze and interpret social media data, including text data. By detecting linguistic markers of distress and other risk factors, these systems can help identify individuals with a risk of suicidality and provide early interventions to prevent such incidents \cite{vioules2018detection}. 

Several studies indicated the impact of social network reciprocal connectivity on users’ suicidal ideation. Hsiung et al. \cite{hsiung2007suicide} analyzed the changes in user behavior following a suicide case that occurred within a social media group. Jashinsky et al. \cite{jashinsky2014tracking} highlighted the geographic correlation between suicide mortality rates and the occurrence of risk factors in tweets. Colombo et al. \cite{colombo2016analysing} focused on analyzing tweets that contained suicidal ideation, with a particular emphasis on the users' behavior within social network interactions that resulted in strong and reciprocal connectivity, leading to strengthened bonds between users.
NLP techniques, therefore, offer a promising avenue for suicide prevention efforts, enabling more proactive and effective interventions to support those in need.

\vspace{5pt}
\noindent\textbf{Generative Language models:} Ghanadian et al.\cite{ghanadian2023chatgpt} utilized ChatGPT for assessing suicidality from social media posts. They performed Zero-Shot and Few-Shot experiments and extensive performance comparison between ChatGPT and two fine-tuned transformer-based models. They also investigated the impact of different temperature parameters on ChatGPT's response generation. The findings of this paper suggest that ChatGPT achieves notable accuracy in the suicidal risk assessment task; however, transformer-based pre-trained models fine-tuned on human-annotated datasets exhibit superior performance. Furthermore, the analysis provides insights into adjusting ChatGPT's hyperparameters to enhance its effectiveness in assisting mental health professionals with this critical task.

Yang et al.\cite{yang2023evaluations} conducted a comprehensive evaluation of ChatGPT's mental health analysis and emotional reasoning ability across five tasks. They also investigated the impact of different emotion-based prompting strategies. Additionally, they explored the use of generative models to generate explanations for the decisions made by ChatGPT, aiming for interpretable mental health analysis. The experimental results revealed that ChatGPT performed better than traditional neural network-based methods such as Convolutional Neural Network (CNN) and Gated Recurrent Unit (GRU) in mental health analysis but still lagged behind advanced task-specific methods. 

\vspace{5pt}
\noindent\textbf{Available Dataset:} Several datasets have been collected from social media platforms to serve as a resource for creating suicidal ideation detection systems. These datasets encompass a wide range of information collected from various social media sources, including \textit{Twitter}, \textit{Reddit} and other user-generated content. Sinha et al.\cite{sinha2019suicidal} created a manually annotated dataset from \textit{Twitter} using a lexicon of suicidal phrases and a lexicon along with the social engagement data associated with real-time and historical tweets. The resulting dataset consists of 34,306 tweets with two labels, \textit{Suicidal} and \textit{Non-Suicidal}. Gaur et al.\cite{gaur2019knowledge} collected and annotated a 5-label Suicide Risk Severity Assessment dataset from Reddit, which includes \textit{Suicidal Ideation (ID)}, \textit{Suicidal Behavior (BR)}, \textit{Actual Attempt (AT)}, \textit{Suicide Indicator (IN)} and \textit{Supportive (SU)} categories. This dataset is extracted from \textit{SucideWatch}\footnote{\href{https://www.reddit.com/r/SuicideWatch/}{SuicideWatch subreddit}} subreddit and has been annotated by four practicing clinical psychiatrists, ensuring the accuracy and reliability of the annotations. The dataset comprises a total of 500 posts, which have been carefully selected to represent a diverse range of content related to suicidal ideation.

Another widely referenced dataset in the field of suicidal ideation detection is the University of Maryland Reddit Suicidality Dataset(UMD) \cite{zirikly2019clpsych,shing2018expert}. The UMD dataset is a collection of Reddit posts and comments created by individuals who expressed suicidal thoughts or behaviors. The dataset contains over 100,000 posts and comments collected from various subreddits, including those related to mental health and suicide prevention, such as \textit{Depression}\footnote{\href{https://www.reddit.com/r/depression/}{Depression subreddit}} and \textit{SucideWatch} subreddits. The data was collected over a period of several years and includes the content of the posts and comments, as well as the location and timing of the posts.
 
\subsection{Synthetic Data collection} 
To overcome the limitations of real-world data availability, NLP researchers have explored the use of synthetic datasets for several applications. 
For example, He et al.\cite{he2022generate} utilized language models to generate synthetic unlabeled text. They introduced the Generate, Annotate, and Learn (GAL) framework that leverages synthetic text for knowledge distillation, self-training, and few-shot learning purposes. To generate the data, they fine-tune pre-trained language models on relevant datasets with limited examples. The synthetic text is then annotated with soft pseudo labels using the best available classifier for knowledge distillation and self-training. This paper achieves state-of-the-art results for knowledge distillation with 6-layer transformers on the GLUE leaderboard\cite{rashid2021mate}.

Bonifacia et al.\cite{bonifacio2022inpars} presents an effective approach to leverage LLMs in retrieval tasks, resulting in significant improvements across various datasets. Instead of directly utilizing LLMs during the retrieval process, they harness the LLMs' capabilities to generate labeled data using a few-shot learning approach. Subsequently, they fine-tune smaller retrieval models on this synthetic dataset and employ them to re-rank the search results obtained from a primary retrieval system. They provide a novel method to adapt LLMs for Information Retrieval (IR) tasks that were previously deemed infeasible due to their demanding computational requirements. By shifting the computational burden from the retrieval stage to the generation of synthetic data for training, they make it feasible to exploit the power of LLMs without compromising efficiency. In an unsupervised setting, their approach significantly outperforms recently proposed methods, highlighting its superiority in terms of retrieval performance and scalability.

\section{Methodology}
\label{sec:methodology}

\begin{figure}[t]
\centering
\includegraphics[width=\columnwidth]{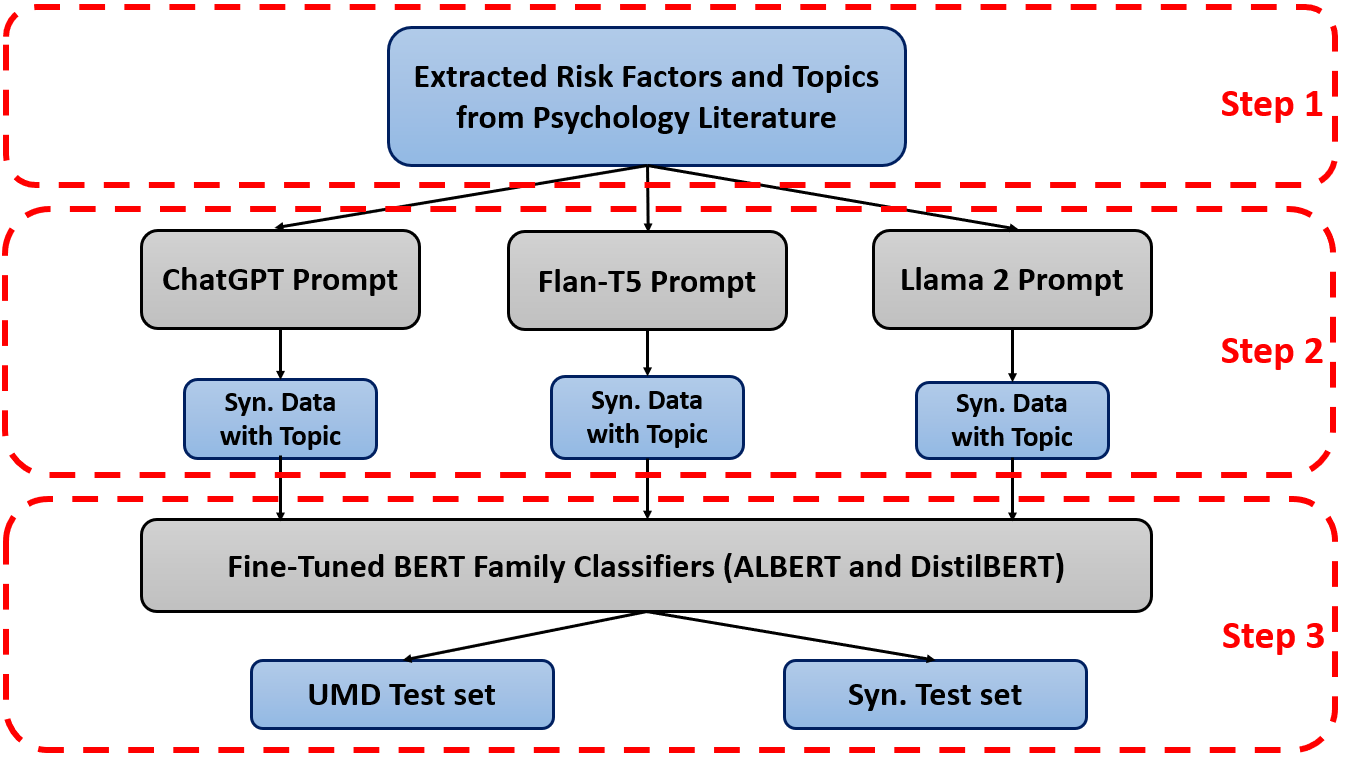}
\caption{Workflow of the proposed methodology}
\label{fig:workflowdiagram}
\end{figure}

In this section, we elaborate on our proposed methodology. Figure \ref{fig:workflowdiagram} shows the workflow we use to generate synthetic datasets and our testing process. As shown in this figure, our method has three steps:

\begin{itemize}

    \item \textbf{STEP 1- Domain knowledge Extraction:} Extract relevant social factors from the psychology literature for an informed prompting of GLLMs in data synthesis. 

    \item \textbf{STEP 2- Synthetic Data Generation:} Use three GLLMs to generate socially aware synthetic data, that is, data that covers a wide range of suicide-related topics. 
    
    \item \textbf{STEP 3- Evaluate the effectiveness of Synthetic data:} Train state-of-the-art classifiers with real-world, synthetic, and augmented datasets and test those classifiers on real-world as well as synthetic test sets.
   
\end{itemize}

In the following, we explain the three steps described above. The complete implementation of our project, including Zero-Shot Learning and Few-Shot Learning of GLLMs, as well as the fine-tuned classifiers, is available on GitHub \footnote{\url{https://github.com/Hamideh-ghanadian/Synthetic_Data_Generation_using_Generative_LLMs}}.

\subsection{Suicide Related Topics in Psychology }
\label{sec:topics}
We conducted a comprehensive search across various academic databases, including PsycINFO, PubMed, and Google Scholar, with keywords and combinations such as ``suicide'', ``suicidal ideation'' and ``psychology'' to identify relevant articles, research papers, and review papers.
Thematic analysis was employed to identify the most recurring and significant topics across the included studies. Repeated topics that demonstrate relevance to suicide in psychology were considered the most related topics.

Based on our analysis of the literature, the following social and psychological factors were consistently reported in relation to suicidal ideation in psychology. These topics are not listed in a specific order of importance but represent the consistently reported themes in the literature reviewed:

\vspace{5pt}
\noindent\textbf{Depression:}
Depression emerged as a frequently reported topic, highlighting its strong association with suicidal ideation. Numerous studies have explored the relationship between depressive symptoms, including sadness, loss of interest, feelings of worthlessness, and the increased risk of suicidal thoughts.\cite{boergers1998reasons,klonsky2016suicide, vilhjalmsson1998factors}

\vspace{5pt}
\noindent\textbf{Anxiety:}
Anxiety disorders were also commonly associated with suicidal ideation. Research has emphasized the link between excessive worry, fear, and agitation and the presence of suicidal thoughts and behaviors\cite{lazaro2023predictive}.

\vspace{5pt}
\noindent\textbf{Unemployment:}
The experience of unemployment has been consistently identified as a topic closely related to suicidal ideation. Studies have examined the psychological distress and negative impact on self-esteem and social support that can arise from unemployment, contributing to suicidal ideation \cite{lee2010prevalence}.

\vspace{5pt}
\noindent\textbf{Hopelessness:}
Hopelessness, characterized by a lack of optimism and a perceived absence of future prospects, has been consistently linked to suicidal ideation. Studies have demonstrated the significant role of hopelessness as a predictor of suicidal thoughts \cite{boergers1998reasons,klonsky2016suicide,lee2010prevalence}.

\vspace{5pt}
\noindent\textbf{Anger:}
The expression and experience of anger have been reported as influential factors in suicidal ideation. Unresolved anger, hostility, and intense emotional distress have been associated with an increased risk of suicidal thoughts\cite{boergers1998reasons}.

\vspace{5pt}
\noindent\textbf{Perfectionism:}
Perfectionism, marked by excessively high standards and self-criticism, has been identified as a psychological factor related to suicidal ideation. Research has explored the relationship between perfectionistic tendencies and the development of suicidal thoughts and behaviors \cite{boergers1998reasons}.

\vspace{5pt}
\noindent\textbf{Family Issues:}
Family-related issues, such as conflict, dysfunctional dynamics, and poor communication, have consistently emerged as topics associated with suicidal ideation. These factors can contribute to a sense of isolation, distress, and feelings of being a burden, increasing the risk of suicidal thoughts\cite{lee2010prevalence, vilhjalmsson1998factors}.

\vspace{5pt}
\noindent\textbf{Relationship Problems:}
Difficulties in intimate relationships, including conflicts, breakups, and marital dissatisfaction, have been reported as significant topics in relation to suicidal ideation. Relationship problems can contribute to emotional distress and feelings of hopelessness, leading to thoughts of suicide \cite{lee2010prevalence}.

\vspace{5pt}
\noindent\textbf{Financial Crisis:}
Financial difficulties and crises have been consistently linked to suicidal ideation. Economic stressors, such as debt, unemployment, and financial insecurity, can contribute to psychological distress and an increased risk of suicidal thoughts \cite{lazaro2023predictive}.

\vspace{5pt}
\noindent\textbf{Education:}
Issues related to educational pressures, academic stress, and performance expectations have been reported as topics associated with suicidal ideation. Research has highlighted the impact of academic-related stressors on mental well-being and the risk of suicidal thoughts among students \cite{farabaugh2012depression}.

\vspace{5pt}
\noindent\textbf{Bullying:}
Bullying, including physical, verbal, or cyber-bullying, has consistently emerged as a significant topic related to suicidal ideation. The experience of bullying can lead to social isolation, low self-esteem, and emotional distress, contributing to the development of suicidal thoughts \cite{lazaro2023predictive}.

\vspace{5pt}
\noindent\textbf{Death of Loved Ones:}
The loss of close family members or friends through death has been reported as a topic associated with suicidal ideation. Grief, feelings of loneliness, and a sense of being unable to cope with the loss can increase the risk of suicidal thoughts \cite{peteet2010unimaginable}.

\vspace{5pt}
\noindent\textbf{Immigration:}
Issues related to immigration, discrimination, and racism have been identified as topics linked to suicidal ideation. Experiences of marginalization, social exclusion, and acculturative stress can contribute to psychological distress and suicidal thoughts among individuals facing these challenges \cite{ratkowska2013suicide, hovey2000acculturative}.

\vspace{5pt}
\noindent\textbf{Racism:}
Studies have consistently highlighted the significant impact of racial discrimination on suicidal ideation. Experiencing racism and racial prejudice can increase the risk of suicidal thoughts.\cite{keum2023gendered}

\subsection{Synthetic Data Generation}
We utilized three generative language models to generate a synthetic dataset related to suicidal ideation. GLLM's foundation is constructed with transformers. Transformers are a class of deep learning models, first introduced by Vaswani et al.\cite{vaswani2017attention} in 2017.
Researchers build state-of-the-art NLP models using transformer-based architectures because they can be quickly trained on large datasets, and studies have shown that they are better at modeling long-term dependencies in natural language text \cite{wolf2020transformers}. GLLMs, including ChatGPT, FlanT5, and Llama are designed with the primary purpose of generating coherent and contextually relevant text. They excel at tasks such as text generation \cite{luo2022biogpt},  completion \cite{xie2022discrimination}, and dialogue generation\cite{mi2022pangu}. These models are typically based on decoder transformer architectures and focus on the generative aspect of language which involves the auto-regressive generation, where the models predict the next word based on the preceding context. Generative models are trained on a vast corpus of text, however, their main strength lies in their ability to generate text that flows naturally and contextually appropriate.

We aim to build a diverse dataset in order to train a generalizable and robust model in suicidal ideation detection. In total, nine different datasets are generated with different specifications and models.

\subsubsection{ChatGPT} The language model utilized by ChatGPT is \textit{gpt-3.5-turbo}\footnote{\url{https://platform.openai.com/docs/models/gpt-3-5}}, which is one of the most advanced language models developed by OpenAI. 
ChatGPT accept a sequence of messages as an input and produce a message generated by the model as an output. Although the chat format is primarily intended for conversations spanning multiple turns, it is also equally useful for single-turn tasks that do not involve any conversations. We used the \textit{OpenAI Python library}\footnote{\url{https://github.com/openai/openai-python}} to access the \textit{ChatCompletion} functionality of the \textit{gpt-3.5-turbo} model through its API. 

In this project, we evaluate the capability of ChatGPT in Zero-Shot Learning and Few-Shot Learning settings to generate a diverse suicidality dataset. However, we are primarily focused on Zero-Shot Learning methods as ChatGPT has exhibited superior performance in this setting compared to Few-Shot Learning for a suicidal ideation detection task. Ghanadian et al. \cite{ghanadian2023chatgpt} conducted an extensive comparison of the Zero-Shot and Few-Shot approaches using ChatGPT. According to their findings, fine-tuning a model on Few-Shot setting might yield poorer performance compared to Zero-Shot in various scenarios. In Few-Shot, with few examples available for fine-tuning, the risk of overfitting increases. The model might learn specific nuances or noise within the limited few-shot data, leading to poor generalization on unseen examples. Moreover, Few-shot learning relies on a small subset of labeled examples, which might not adequately represent the entire diversity of the dataset. The model might fail to capture the complexity and variability present in the broader dataset during fine-tuning. 

The temperature hyperparameter in ChatGPT is a crucial parameter that influences the generated output. A higher temperature value, such as 1.0, increases the randomness and produces more varied responses. Conversely, a lower temperature value, such as 0.1, reduces randomness and generates more focused and deterministic responses. Ghanadian et al.\cite{ghanadian2023chatgpt} investigated the effect of the temperature parameter on the generated output of ChatGPT for suicide risk assessment. Furthermore, the authors introduced a parameter known as the "Inconclusiveness Rate," which indicates the proportion of test cases that do not produce a definitive or conclusive result. According to their paper, this parameter decreases as the temperature parameter is increased. As such, for this project, we have configured the temperature parameter of ChatGPT to be 1. 

We generated five datasets using ChatGPT. Four of these datasets are informed by 14 main suicide-related topics in psychology, while one dataset is generated without providing any specific topics. For incorporating suicide-related topics in data generation, we utilized prompt engineering techniques. 
Prompt engineering involves carefully crafting and designing the prompts provided to the model to elicit desired responses. By employing prompt engineering strategies, we aimed to enhance the quality and relevance of the generated dataset. This methodology allowed us to tailor the dataset generation process to align with our specific objectives and requirements. For prompt engineering, we drew inspiration from a short course on ChatGPT Prompt Engineering\footnote{\href{https://www.deeplearning.ai/short-courses/chatgpt-prompt-engineering-for-developers/}{ChatGPT Prompt Engineering for Developers}}. 

In both Zero-Shot and Few-Shot Learning, we initiated the prompt construction process with a simple initial prompt and iteratively refined it through multiple rounds of trial and error. This iterative approach allowed us to gradually evolve the prompt, making necessary adjustments based on the observed outcomes. The following example shows a prompt we used to generate a binary dataset (\textit{Suicide/non-suicide}), guided by the topics introduced in Section \ref{sec:topics}, in Zero-shot Learning setting. 

\begin{small}
\begin{lstlisting}[language=Python,breaklines=true] 
Criteria=""" These are the criteria of different suicide risk level:
    Risk Level=Non Suicidal: I do not see evidence that this person is at risk for suicide
    Risk Level=Suicidal: I believe this person is at high risk of attempting suicide in the near future. """
   
prompt = f"""
Your task is to generate a suicidal text for each of the following "topics" with different Risk levels.

    1-Depression
    2-Anxiety
    3-Hopelessness
    4-Anger
    5-Perfectionism
    6-Family issues
    7-Relationship problems
    8-Unemployment
    9-Financial Crisis
    10-Education 
    11-Being Bullied
    12-Death of closed one
    13-Immigration
    14-Racism

Provide the answers in JSON format with the following columns: text, topic, risk level.

Risk level criteria: ```{Criteria}```

"""        
\end{lstlisting}
\end{small}

In Few-Shot Learning, the prompt is structured to include two examples for each category (8 in total) from the training set of UMD Dataset, followed by a text generation question. This approach enables the model to learn from a limited set of labeled examples before generating a dataset. Moreover, by combining the Few-Shot Learning methodology with the inclusion of psychology topics in the prompt, we aim to enhance the model's ability to generate meaningful and contextually relevant responses when dealing with suicide-related discussions.

\subsubsection{Flan-T5} FLAN-T5 models are instruction fine-tuned across a diverse set of tasks, aiming to enhance their zero-shot performance on various tasks. During instruction tuning, pretrained models undergo fine-tuning using drafts of instructions that guide them on how to perform a specific task. These instructions can include real-time feedback to assist the model in learning from its mistakes and improving at a faster rate. By providing explicit guidance and incorporating feedback mechanisms, the instruction-tuning process enables the model to refine its performance and enhance its ability to accurately execute the given task. This iterative approach of incorporating instructions and feedback facilitates the model's learning process, allowing it to adapt and improve its performance based on the provided guidance.

In this project, we utilized \textit{Flan-T5-XXL}\footnote{\url{https://huggingface.co/google/flan-t5-xxl}} presented by Google Research \cite{chung2022scaling} in a Zero-Shot setting. Two datasets are generated using Flan-T5, one with topics and another without topics. Moreover, similar to ChatGPT, the temperature value is set to 1, and the same prompt structure is utilized.  

\subsubsection{Llama 2}
LLaMA (Large Language Model Meta AI) is an auto-regressive language model constructed based on transformer architecture. Similar to other generative models, LLaMA operates by taking a sequence of words as its input and making predictions about the subsequent word, iteratively producing text in a recursive manner.
It is a collection of state-of-the-art foundational language models, with parameter counts ranging from 7 billion to 65 billion. The foundation models were trained on large unlabeled datasets, making them ideal for fine-tuning on a variety of tasks. The newest version of this model, Llama 2, expanded its pre-training corpus size, allowing the model to learn from a more extensive and diverse set of publicly available data. Additionally, the context length of Llama 2 has been doubled, enabling the model to consider a more extensive context when generating responses, leading to improved output quality and accuracy \cite{touvron2023llama}. In this paper, we used Llama 2-13B, presented by Meta in the Zero-Shot setting. In total, we generated two datasets with Llama2, one with topics and another without topics. These datasets were created using the temperature of 1 and maintained the same prompt structure as ChatGPT.


\subsection{Evaluation of Synthetic Dataset}\label{sec: finetuned}

To evaluate the utility and effectiveness of synthetic datasets, we fine-tuned pre-trained transformer-based language models, ALBERT and DistilBERT, to train classifiers with each set of the generated synthetic data. We compared the trained classifiers with classifiers with similar structures fine-tuned with real-world data as the benchmark model. 

ALBERT and DistilBERT are two pre-trained language models from the BERT family of LMs. The BERT model was initially proposed by Delvin et al.\cite{devlin2018bert} as a bidirectional language model pretrained on a large corpus comprising the Toronto Book Corpus and Wikipedia. The model is named bidirectional because it can simultaneously gather the context of a word from either direction. Unlike the generative models such as ChatGPT, FlanT5 or Llama, which include a decoder structure, the BERT family of language models are encoder models and can be fine-tuned for specific tasks such as classification tasks. 

The ALBERT model was proposed by Lan et al.\cite{lan2019ALBERT} to reduce memory consumption and increase the training speed compared to BERT. In other words, ALBERT is a more lightweight version of BERT that maintains its high level of accuracy, making it a powerful tool for various NLP applications.
The DistilBERT model was proposed by Sanh et al. \cite{sanh2019distilbert}. The authors reported it has 40\% fewer parameters than BERT and runs 60\% faster while preserving over 95\% of BERT’s performances as measured on the GLUE language understanding benchmark. Both models are designed as lightweight alternatives to BERT, with ALBERT emphasizing parameter efficiency and DistilBERT focusing on knowledge transfer through distillation. Overall, ALBERT, with a smaller number of parameters, shows more efficient performance compared to DistilBERT.

To fine-tune these models, we utilized the Huggingface library \cite{wolf2019huggingface}. The Huggingface is an open-source library and data science platform that provides tools to build, train and deploy ML models. We compare our classification results with baseline ALBERT\footnote{\href{https://huggingface.co/albert-base-v2}{ALBERT}} and DistilBERT\footnote{\href{https://huggingface.co/distilbert-base-uncased}{DistilBERT}} models fine-tuned on the UMD dataset by Ghanadian et al. \cite{ghanadian2023chatgpt}. We used the Trainer\footnote{\href{https://huggingface.co/docs/transformers/main_classes/trainer}{Trainer}} class from Huggingface transformers\footnote{\href{https://huggingface.co/docs/transformers/index}{Huggingface Transformers}} for feature-complete training in PyTorch.

The hyperparameters were selected based on the default values commonly used in similar studies. The final hyperparameters used in our experiments were Learning Rate= $2e^{-5}$, Batch Size = 4, Dropout Rate = 0.1, and Maximum Sequence Length = 512. By comparing the performance of these models on synthetic datasets against the baseline, we can assess the efficiency of using the synthetic datasets and gauge the improvements achieved through our fine-tuning process.

To conduct a comprehensive assessment of the fine-tuned classifiers' performance, we generated two distinct sets of testing subsets. Furthermore, we created an augmented dataset to showcase the application of synthetic data in the suicidal ideation detection domain.

\subsubsection{Testing subsets}\label{subsubsec: datatest}
We selected two test sets for evaluation purposes:
The first test dataset is the test subset of the UMD datasets utilized in  \cite{ghanadian2023chatgpt}, which is annotated as a 4-class dataset. We employed a 10-20-70 split for validation, test, and training sets, respectively. Out of the entire dataset, 10\% was allocated for validation purposes, ensuring the model's hyperparameters and configurations were appropriately set. 20\% of the data was set aside as a test set to evaluate the model's performance on unseen data and ensure its generalizability. The remaining 70\% formed the training set, where the bulk of the data was utilized to train the model and learn the underlying patterns. This distribution was chosen to provide substantial data for training while reserving enough distinct data for validation and robust performance testing. The detailed description of the Multi-class UMD dataset is presented in Table~\ref{tab:UMD}.
\begin{table}[ht]
\centering
\caption{The description of the training and testing subset of UMD Dataset used in \cite{ghanadian2023chatgpt} for multi-class task}
\label{tab:UMD}
\huge
\resizebox{\columnwidth}{!}{%
\begin{tabular}{c|cccc}
Multi-class Dataset              & No Risk & Low Risk & Moderate Risk & High Risk \\ \hline
Training Subset    &27.45 \%     & 16.39 \%      & 31.90 \%           & 24.24 \%       \\ \hline
Number of Users    & 154     & 92      & 179           & 136       \\ \hline\hline
Testing Subset    & 24.41 \%     & 11.62 \%      & 26.74 \%           & 37.20 \%       \\ \hline
Number of Users    & 42     & 20      & 46           & 64       \\ \hline
\end{tabular}%
}
\end{table}

Furthermore, to employ binary classification, we binarize the UMD Dataset. Based on the definition of each class, ``\textit{No Risk}'' and ``\textit{Low Risk}'' classes are considered as Non-Suicidal and ``\textit{Moderate Risk}'' and``\textit{High Risk}'' as Suicidal. Table~\ref{tab: binary dataset} presents the description of binarized UMD dataset. 
\begin{table}[ht]
\centering
\tiny
\caption{The description of the training and testing subset of UMD Dataset for binary task}
\label{tab: binary dataset}
\resizebox{0.4\textwidth}{!}{%
\begin{tabular}{c|clcl}
Binary Dataset     & \multicolumn{2}{c}{Non Suicidal} & \multicolumn{2}{c}{Suicidal} \\ \hline
Training Subset & \multicolumn{2}{c}{43.84\%}      & \multicolumn{2}{c}{56.14\%}  \\ \hline
Number of Users     & \multicolumn{2}{c}{246}          & \multicolumn{2}{c}{315}      \\ \hline\hline
Testing Subset  & \multicolumn{2}{c}{36.3}         & \multicolumn{2}{c}{63.94}    \\ \hline
Number of Users     & \multicolumn{2}{c}{62}           & \multicolumn{2}{c}{110}      \\ \hline
\end{tabular}
}
\end{table}

The second testing set is composed of 10\% of each synthetic dataset generated in our project. This test set is annotated independently by two human annotators. A notable 89\% of the labels, initially generated by the generative models, were agreed upon by the human annotators. However, for the remaining 11\% of the data, the labels were altered based on the decision of the annotators. In cases where both annotators agreed on a label, that label was retained. Conversely, when disagreements arose, the annotators engaged in discussions to ultimately reach a consensus on the appropriate label. The details of the generated synthetic dataset are presented in Table ~\ref{tab:datasets} as the eleventh dataset.

\subsubsection{Augmented Dataset}\label{subsec: Augment method}
Data augmentation involves enriching a dataset by introducing variations to its existing instances or generating entirely new instances. This process is designed to enhance the diversity and quality of the dataset, which, in turn, can lead to improved model performance and generalization. Hence, in this study, we augment the best performing synthetic dataset generated by LLMs with different subset sizes of UMD dataset. Starting with 10\% of the UMD training subset, this subset is combined with the selected synthetic dataset. The augmented dataset, which is a mix of synthetic and real instances, is used to fine-tune the pretrained models. We continue this process by increasing the number of real data instances, such as 20\% and 30\%, until achieving comparable results to those obtained from the model trained on the full UMD dataset.

\section{Results}\label{sec:Results}
In this section, first, we present the characteristics of each synthetic dataset. Second, we report a comprehensive comparison of the models fine-tuned with them. 
Third, we report the data augmentation results. For evaluation, we report two widely-used metrics in this task, accuracy and F-score, to provide a complete and informative evaluation of the performance of the classification models \cite{sokolova2009systematic}.

\subsection{Synthetic Data Generation}
A total of nine datasets are generated. An extensive description of these datasets, as well as a mixed set and a test subset, is presented in Table~\ref{tab:datasets}. As shown in Table~\ref{tab:datasets}, we created binary datasets and four-class datasets, each with the option of including or not including the topics. 
The binary datasets contain two classes, which allows us to evaluate the model's ability to distinguish between suicidal ideation and non-suicidal instances. On the other hand, the four-class datasets involve multiple categories, enabling us to explore more nuanced predictions of suicidal ideation levels, including ``\textit{No Risk}'', ``\textit{Low Risk}'', ``\textit{Moderate Risk}'' and ``\textit{High Risk}'' classes. Moreover, the option to include or not include the topics in these datasets allows us to investigate the impact of information provided by topics on the model's performance.

\begin{table*}[h]
\centering
\caption{Detailed description of generated synthetic datasets}
\label{tab:datasets}
\resizebox{1.8\columnwidth}{!}{%
\begin{tabular}{cccccc}
\hline
Dataset \# & Model & Learning Method & Topic-Oriented & \# of Class & \# of Instances \\ \hline
1 & Chat GPT & Zero-Shot & Yes & 2 & 549 \\
2 & Chat GPT & Zero-Shot & No  & 2 & 646 \\
3 & Chat GPT & Few-Shot  & Yes & 2 & 545 \\
4 & Chat GPT & Zero-Shot & Yes & 4 & 492 \\
5 & Chat GPT & Few-Shot  & Yes & 4 & 594 \\
6 & Flan-T5  & Zero-Shot & Yes & 2 & 561 \\
7 & Flan-T5  & Zero-Shot & No & 2 & 502 \\
8 & Llama 2  & Zero-Shot & Yes & 2 & 395   \\ 
9 & Llama 2  & Zero-Shot & No & 2 & 613  \\
10 & Mix Dataset  & Zero-Shot & Yes & 2 & 1352  \\
11 & Synthetic Testing Set  & Zero-Shot & N/A & 2 & 318  \\\hline
\end{tabular}%
}
\end{table*}

By comparing the results from datasets with and without topics, we can gain insights into how incorporating topic-related data enhances or influences the model's effectiveness in suicidal ideation detection.
Furthermore, as explained in Section \ref{subsubsec: datatest}, we created a synthetic testing dataset comprising 10\% of each dataset which is annotated by human experts.

\subsection{Fine-Tuned Classifiers}
Two models, ALBERT and DistilBERT are fine-tuned with the generated synthetic datasets. Table~\ref{tab: Multiclasss GhatGPT} presents the results of the performance evaluation of models fine-tuned with multi-class synthetic datasets generated by ChatGPT in Zero-Shot and Few-Shot settings, tested on the multi-class UMD test set. Considering the poor performance of the multi-class synthetic dataset, we have chosen to disregard the multi-class aspect and proceed solely with the binary approach.

\begin{table}[ht]
\centering
\caption{Performance evaluation of ALBERT and DistilBERT models on Multi-class datasets generated by ChatGPT }
\label{tab: Multiclasss GhatGPT}
\huge
\resizebox{0.9\columnwidth}{!}{%
\begin{tabular}{c|c|ccc}
Models                                                 & Metrics  & \multicolumn{1}{c}{\begin{tabular}[c]{@{}c@{}}Non-Synthetic \\ UMD Dataset\end{tabular}} & \multicolumn{1}{c}{\begin{tabular}[c]{@{}c@{}}ChatGPT\\ Zero-Shot\end{tabular}} & {\begin{tabular}[c]{@{}c@{}}ChatGPT\\ Few-Shot\end{tabular}} \\ \hline
\multicolumn{1}{c|}{\multirow{2}{*}{ALBERT}}     & Accuracy & 0.865                                                                                     & 0.41                                                                             & 0.36                                                       \\
\multicolumn{1}{c|}{}                                     & F1-Score & 0.87                                                                                     & 0.43                                                                             & 0.27                                                       \\ \hline
\multicolumn{1}{c|}{\multirow{2}{*}{DistilBERT}} & Accuracy & 0.77                                                                                      & 0.06                                                                             & 0.06                                                      \\
\multicolumn{1}{c|}{}                                     & F1-Score & 0.75                                                                                     & 0.1                                                                              & 0.12                                                       \\ \hline
\end{tabular}
}
\end{table}

Moreover, Table~\ref{tab: Performance} provides the performance of the models trained on the binary synthetic datasets generated by ChatGPT, Flan-T5 and Llama models and tested on the binary UMD testing subset. We compare the results for the synthetic datasets with those of the UMD training set. Table~\ref{tab: Performance} shows that incorporating topic in generating the datasets significantly improves the performance of the models. For instance, for Llama 2, the topic-oriented dataset increased the F1-score and accuracy of the ALBERT model by 10\% and 14\% points, respectively. We also created a mixed dataset, including all topic-oriented datasets, to further evaluate the effects of topics on the performance of the models. With both ALBERT and DistilBERT, an F1-score of 0.82 is achieved by the mixed dataset, which is significantly higher than the DistilBERT model trained on the UMD dataset and comparable with the performance of the ALBERT model fine-tuned on the UMD dataset with an F1-score of 0.87.

Table~\ref{tab: Performance on testing} presents the results of models included in Table~\ref{tab: Performance} but tested on synthetic testing datasets. Similar to the results of Table~\ref{tab: Performance}, all of the topic-oriented datasets show significant improvement compared to the datasets without any topics. ChatGPT-generated training data, with an F1-score of 0.82, exhibits the best performance, while the performances of Flan-T5 and Llama2-generated datasets are acceptable. Moreover, the mixed dataset shows a 0.81 F1-score, which is an 11\% improvement compared to the model trained with the UMD dataset. 
\begin{table*}[]
\centering
\caption{Performance evaluation of the ALBERT and DistilBERT models fine-tuned with binary datasets and tested on UMD testing subset}
\label{tab: Performance}
\resizebox{2\columnwidth}{!}{%
\begin{tabular}{cc|c|ccccc|cc|c}
\textbf{} &
  \textbf{} &
  Non-Synthetic &
  \multicolumn{3}{c}{ChatGPT} &
  \multicolumn{2}{c|}{Flan-T5} &
  \multicolumn{2}{c|}{Llama 2} &
  \multicolumn{1}{l}{Mix Dataset} \\ \cline{3-11} 
\multicolumn{1}{c|}{Models} &
  Metrics &
  UMD Dataset &
  \multicolumn{1}{l}{\begin{tabular}[c]{@{}l@{}}With Topic\\ Few-Shot\end{tabular}} &
  {\begin{tabular}[c]{@{}c@{}}Without Topic\\ Zero-Shot\end{tabular}} &
  \multicolumn{1}{c|}{\begin{tabular}[c]{@{}c@{}}With Topic \\ Zero-Shot\end{tabular}} &
  \multicolumn{1}{l}{Without Topic} &
  With Topic &
  \multicolumn{1}{l}{Without Topic} &
  With Topic &
  With Topic \\ \hline
\multicolumn{1}{c|}{\multirow{2}{*}{ALBERT}} &
  Accuracy &
  0.87 &
  0.67 &
  0.70 &
  \multicolumn{1}{c|}{0.71} &
  0.48 &
  0.62 &
  0.33 &
  0.75 &
  0.77 \\
\multicolumn{1}{c|}{} &
  F1-Score &
  0.87 &
  0.66 &
  0.79 &
  \multicolumn{1}{c|}{0.79} &
  0.54 &
  0.64 &
  0.49 &
  0.78 &
  0.82 \\ \hline
\multicolumn{1}{c|}{\multirow{2}{*}{DistilBERT}} &
  Accuracy &
  0.77 &
  0.61 &
  0.63 &
  \multicolumn{1}{c|}{0.64} &
  0.59 &
  0.77 &
  0.32 &
  0.75 &
  0.76 \\
\multicolumn{1}{c|}{} &
  F1-Score &
  0.75 &
  0.59 &
  0.69 &
  \multicolumn{1}{c|}{0.71} &
  0.61 &
  0.84 &
  0.15 &
  0.77 &
  0.82 \\ \hline
\end{tabular}%
}
\end{table*}

\begin{table*}[]
\centering
\caption{Performance evaluation of the ALBERT and DistilBERT models fine-tuned with binary datasets and tested on synthetic testing subset}
\label{tab: Performance on testing}
\resizebox{2\columnwidth}{!}{%
\begin{tabular}{cc|c|ccccc|cc|c}
\textbf{}                                                  & \textbf{}         & Non-Synthetic & \multicolumn{3}{c}{ChatGPT}                                                                                                                                                                                                                                    & \multicolumn{2}{c|}{Flan-T5}                            & \multicolumn{2}{c|}{Llama 2}                            & \multicolumn{1}{l}{Mix Dataset} \\ \cline{3-11} 
\multicolumn{1}{c|}{Models}                       & Metrics  & UMD Dataset   & \multicolumn{1}{l}{\begin{tabular}[c]{@{}l@{}}With Topic\\ Few-Shot\end{tabular}} &{\begin{tabular}[c]{@{}c@{}}Without Topic\\ Zero-Shot\end{tabular}} & \multicolumn{1}{c|}{\begin{tabular}[c]{@{}c@{}}With Topic \\ Zero-Shot\end{tabular}} & \multicolumn{1}{l}{Without Topic} & With Topic & \multicolumn{1}{l}{Without Topic} & With Topic & With Topic                      \\ \hline
\multicolumn{1}{c|}{}                                      & Accuracy & 0.67          & 0.71                                                                              & 0.81                                       & \multicolumn{1}{c|}{0.81}                                     & 0.34                              & 0.63       & 0.48                              & 0.70       & 0.83                            \\
\multicolumn{1}{c|}{\multirow{-2}{*}{ALBERT}}     & F1-Score & 0.70          & 0.69                                                                              &  0.78                                       & \multicolumn{1}{c|}{0.82}                                     & 0.41                              & 0.69       & 0.24                              & 0.73       & 0.81                            \\ \hline
\multicolumn{1}{c|}{}                                      & Accuracy & 0.40          & 0.65                                                                              & 0.83                                                              & \multicolumn{1}{c|}{0.85}                                                            & 0.63                              & 0.86       & 0.49                              & 0.63       & 0.78                            \\
\multicolumn{1}{c|}{\multirow{-2}{*}{DistilBERT}} & F1-Score & 0.61          & 0.61                                                                              & 0.81                                                              & \multicolumn{1}{c|}{0.81}                                                            & 0.69                              & 0.84       & 0.12                              & 0.69       & 0.73                            \\ \hline
\end{tabular}%
}
\end{table*}

\subsection{Data Augmentation}
Based on the results presented in Table~\ref{tab: Performance} and Table~\ref{tab: Performance on testing}, the datasets generated by ChatGPT in the Zero-Shot setting show the best results compared to the other datasets. As explained in section \ref{subsec: Augment method}, the augmented dataset now contains a mix of synthetic and real data instances. The augmented dataset is used to fine-tune the pretrained models and then evaluated on two separate testing sets.
In each iteration, three folds, each comprising 10\% of non-overlapping random samples from the UMD dataset, are added to the synthetic data. Subsequently, the average\footnote{we also calculated the standard deviation of the metrics which were always <0.02.} of the accuracy and F1-score are calculated and reported in Table~\ref{tab: Augmentation}. 
If the model's performance with the augmented dataset is less than the model trained with the UMD dataset, additional real-world data is gradually incorporated. For instance, the percentage of real data can be increased to 20\% in the next iteration, and the training and evaluation process is repeated.

\begin{table}[h]
\centering
\caption{Performance evaluation of the ALBERT  model fine-tuned with the augmented dataset (synthetic data + a subset of the UMD train set) and tested on UMD and synthetic testing subsets}
\label{tab: Augmentation}
\resizebox{\columnwidth}{!}{%
\begin{threeparttable}
\Huge
\begin{tabular}{c|c|c|ccc}
Test Set                               & Metric   & {\begin{tabular}[c]{@{}c@{}}UMD \\ Dataset\end{tabular}} & {\begin{tabular}[c]{@{}c@{}}10\% \\ (Avg. of 3 Folds)\tnote{*}\end{tabular}} & {\begin{tabular}[c]{@{}c@{}}20\%  \\ (Avg. of 3 Folds)\tnote{*}\end{tabular}} & f{\begin{tabular}[c]{@{}c@{}}30\% \\ (Avg. of 3 Folds)\tnote{*}\end{tabular}} \\ \hline
\multirow{2}{*}{UMD Testing Set}      & Accuracy & 0.87                                                   & 0.75                                               & 0.81                                                & 0.83                                              \\
                                                & F1-Score & 0.87                                                   & 0.79                                               & 0.84                                                & 0.88                                               \\ \hline
\multirow{2}{*}{Synthetic Testing Set} & Accuracy & 0.67                                                   & 0.87                                               & 0.87                                                & 0.90                                               \\
                                                & F1-Score & 0.70                                                   & 0.83                                               & 0.86                                                & 0.88                                               \\ \hline
\end{tabular}
\begin{tablenotes}
    \item[*] Standard Deviation$<$ 2\%
\end{tablenotes}
\end{threeparttable}
}
\end{table}

Throughout the iterations, the model's performance is closely monitored and compared to the baseline model trained solely on the UMD dataset. The aim is to identify the point at which the augmented dataset starts producing results comparable to or even surpassing those of the baseline model. The process continues until an optimal percentage of real data is found, where the model achieves similar results as the baseline. This ratio indicates the ideal balance between synthetic and real data for achieving high model performance and generalization. Table~\ref{tab: Augmentation} shows the results of each augmentation process until we achieved the F1-score of 0.87 on the UMD testing subset at 30\% augmentation rate and F1-score of 0.85 on the synthetic testing subset at 10\% augmentation rate.

\section{Discussions} \label{sec:Discussion}
This study focuses on the generation of synthetic datasets using generative models and subsequently assessing the performance of models fine-tuned with these datasets. Our synthetic data generation framework addresses two limitations of real-world data collection and annotation. First, we address the data scarcity and annotation cost by generating micropost-like suicidal/non-suicidal text. Second, we address the lack of diversity in real-world data by forcing the generative models to create a balanced number of examples related to each of the psychological and social factors impacting suicidality.   Integrating insights from psychology into the NLP pipeline in this context can illuminate previously unexplored facets of suicide and mental health detection in social media. 

We created several datasets, including binary and multi-class, in Zero-Shot and Few-Shot settings, topic-oriented and non-topic-oriented, with three different generative LLMs. Early in our experiments (Table {~\ref{tab: Multiclasss GhatGPT})}, we observed that ChatGPT is not able to produce high-quality multi-class datasets in either the Zero-Shot or the Few-Shot settings. 
Generating multi-class datasets using LLMs such as ChatGPT is more complex and challenging task due to the inherent complexities involved in distinguishing between multiple and fine-grained, classes. Even with the availability of a high-quality dataset, one should anticipate lower accuracies in multi-class scenarios. This is largely attributed to the ambiguous boundaries that exist between these classes, creating a complex landscape that proves difficult for any classifier to navigate successfully. 
Moreover, the creation of such datasets necessitates not only a detailed prompt but also specific instructions that outline the multi-class scenarios. This process demands a nuanced understanding and a level of specificity that often poses a considerable challenge to ChatGPT. Longer\cite{bang2023multitask}.

As a result, we opted to exclusively create binary datasets and focus our investigation on how topics impact the overall generalizability of the fine-tuned models.

Our results show the critical role of incorporating domain knowledge in synthetic data generation. We extracted the relevant social topics from the Psychology literature and used that to create more focused prompts for data generation. Table~\ref{tab: example} displays a selection of binary samples generated by ChatGPT within the synthetic dataset using social topics. The table provides an illustration of specific examples generated by this GLLM.

\begin{table*}[]
\caption{Generated synthetic samples using extracted social and psychological topics by ChatGPT}
\label{tab: example}
\resizebox{\textwidth}{!}{%
\begin{tabular}{c|c|l}
Topics &
  Labels &
  \multicolumn{1}{c}{Sample} \\ \hline
\multirow{2}{*}{Depression} &
  1 &
  \begin{tabular}[c]{@{}l@{}}Growing up, I always felt like nobody wanted me. No one ever seemed to care about my thoughts, feelings, \\ or emotions. As I got older, these feelings intensified, and I started to feel like I was a burden on everyone \\ around me. I never felt like I belonged anywhere and I constantly struggled with my depression. The constant \\ emptiness and loneliness eventually became too much to bear, and I began to contemplate ending my life.\end{tabular} \\ \cline{2-3} 
 &
  0 &
  \begin{tabular}[c]{@{}l@{}}I've been struggling with mental health problems for as long as I can remember. Depression have been my \\ constant companions. But I've never let it stop me from living my life. I've learned to manage my symptoms \\ with the help of therapy and medication. Some days are harder than others, but I keep pushing forward and \\ focusing on the good in my life. I've come to accept that mental illness is a part of me, but it doesn't define me.\end{tabular} \\ \hline
\multirow{2}{*}{Bullying} &
  1 &
  \begin{tabular}[c]{@{}l@{}}I felt like I was constantly being attacked by my classmates. They called me names, pushed me around, \\ and even vandalized my belongings. The worst part was feeling so alone and  misunderstood. I didn't \\ want to burden my parents with my problems so I kept it all inside. The idea of ending it all seemed like\\  the only way to escape the pain and torment.\end{tabular} \\ \cline{2-3} 
 &
  0 &
  \begin{tabular}[c]{@{}l@{}}When I was in middle school, I was constantly bullied for being overweight. It was tough going to school \\ every day and constantly being made fun of by my peers.  I felt so alone and helpless. But I didn't let the\\  bullies control my life. I joined a support group and found friends who lifted me up. I also started exercising \\ and eating healthier, not to please anyone else, but to feel better about myself. It wasn't an easy journey, but \\ it was worth it. I'm happy to say that today I am confident and proud of who I am.\end{tabular} \\ \hline
\multirow{2}{*}{Unemployment} &
  1 &
  \begin{tabular}[c]{@{}l@{}}After losing my job, I stayed unemployed for months. My savings ran out and with no source of income, \\ my bills piled up. The constant fear of not being able to provide for myself  and my family drove me to \\ the brink of despair. I feel worthless and like a burden on everyone around me. The future seems bleak\\  and hopeless, and I wonder if it's worth it to keep going.\end{tabular} \\ \cline{2-3} 
 &
  0 &
  \begin{tabular}[c]{@{}l@{}}After graduating from college, I struggled to find a job in my field for a few months. It was frustrating \\ and disheartening, but I kept applying and networking. Eventually, I landed a job in a related field that \\ I enjoy. It wasn't my dream job, but it paid the bills and gave me experience. I'm still looking for my\\  dream job, but I'm grateful for what I have and optimistic about my future prospects.\end{tabular} \\ \hline
\multirow{2}{*}{Relationship problems} &
  1 &
  \begin{tabular}[c]{@{}l@{}}I thought I had found the one but it seems like I was wrong because he left me for someone else. I don't \\ know how to deal with this pain . I can't sleep, I can't eat, and I just want to disappear. Maybe everything\\  would be easier if I just ended it all.\end{tabular} \\ \cline{2-3} 
 &
  0 &
  \begin{tabular}[c]{@{}l@{}}My relationship with my partner hasn't been going well lately. We have been arguing over small things, \\ and it's affecting our mental health. We decided to go for couples therapy, and it's been a turning point for\\  us. We learned to communicate better and understand each other's perspective. Now we are in a better \\ place and happier than ever before.\end{tabular} \\ \hline
\end{tabular}%
}
\end{table*}

Table~\ref{tab: Performance} presents the results of fine-tuned models on synthetic datasets and tested on the UMD dataset. Comparison between topic-oriented datasets and no topic-oriented datasets shows the significant effects of including the topics on the performance of the generated 
datasets. Informing the data generation with topics in Flan-T5 and Llama2 increased the F1-Score of the ALBERT model by 10\% and 29\% points, respectively. Fine-tuning models on topic-oriented synthetic datasets allows them to gain diverse domain-specific knowledge and patterns. 
Moreover, non-topic-oriented synthetic datasets might lack specificity, leading to noise and irrelevant content. In contrast, topic-oriented datasets are curated to focus on a specific domain, reducing the chances of generating irrelevant or out-of-context text. 

We showed that the BERT family fine-tuned with real-world data can achieve an F1 score ranging from 0.75 to 0.87, depending on the complexity of their structure. Specifically, DistilBERT, a less efficient model from the BERT Family, achieves an F1-score of 0.75, while ALBERT, a more optimized model designed for speed and accuracy, attains an F1-score of 0.87. In contrast, both DistilBERT and ALBERT achieve a consistent F1-score of 0.82 when trained on purely synthetic data and tested on real-world data. With this, we demonstrate that the diversity of synthetic data compensates for model complexity irrespective of its architecture. This not only underscores the considerable potential of synthetic data but also suggests that it can mitigate the limitations of real-world data in capturing diverse topics. Most notably, our results emphasize an optimal strategy that involves augmenting synthetic data with real data. This innovative method achieves performances comparable to the ALBERT model, even when relying on merely 30\% of the manually annotated dataset. This solidifies our proposed method as a cost-effective alternative, addressing the challenges of data scarcity and diversity more effectively than the current benchmarks. However, 
Synthetic datasets often exhibit a distributional shift from real-world data. This shift arises due to the inherent differences in the data generation processes between synthetic and real domains. As a result, models trained solely on synthetic datasets may not be applicable in real-world situations, leading to a lack of robustness and adaptability. Therefor, exploring hybrid approaches that combine synthetic and real-world data for training can offer a more comprehensive solution. Leveraging both sources allows models to learn from the strengths of synthetic data while adapting to the intricacies of real-world environments.

As presented in Table~\ref{tab: Performance} and Table~\ref{tab: Augmentation}, our study's central objective was to investigate the potential of synthetic and augmented data in training models to perform effectively on real-world data . Given this setup, the chance of overfitting is inherently reduced since the training (synthetic) and testing (real-world) datasets are obtained from distinct distributions.
Moreover, to better understand the performance, robustness, and limitations of the fine-tuned classifiers, we curated an additional test set by manual annotation of a subset of the synthetic data. 
Additional tests ensure that the models do not overfit a particular dataset and can handle a variety of data distributions and scenarios. Table~\ref{tab: Performance on testing} presents the performance results of the fine-tuned models evaluated on the human-annotated synthetic dataset. Notably, the topic-oriented ChatGPT dataset stands out with an F1-score of 0.82, demonstrating its superior performance compared to the other datasets. Specifically, the model trained with the UMD dataset falls short in handling the synthetic test set, presumably because of its less diverse topics.

\section{Conclusion and Future works}
\label{sec:future-works}

The accurate identification of suicidal ideation from textual data holds paramount importance for early intervention and prevention efforts. Natural Language Processing (NLP) techniques have shown promise in this domain, but the scarcity and sensitivity of real suicide-related data pose significant challenges. Gathering and annotating real suicide-related data is a resource-intensive and ethically sensitive process. Synthetic data generation methods, such as text generation models and data augmentation techniques, offer a more cost-effective way to supplement real data. Also, our synthetic datasets offer a potential solution by providing additional social and psychological context in training instances
for models to address the limitations of the existing real data. 

Our data augmentation results 
show that incorporating synthetic data into the training pipeline helps diversify the dataset and enhance model generalization. Real data is often limited in size, leading to over-fitting and reduced model performance. However, by carefully blending synthetic data with real data, we can bolster the model's performance while maintaining a balance between practicality and sensitivity.

Moving forward, exploring the diversity of Language Models (LLMs) stands as an intriguing avenue for future research. Investigating and quantifying the extent of diversity within LLMs across various domains, languages, and training methodologies could offer valuable insights. Future works could delve into developing robust metrics or methodologies specifically tailored to assess and measure diversity within these models.

This paper has effectively highlighted the advantages of using synthetic data generation techniques in detecting suicidal ideation. However, the field still presents numerous opportunities for further research and refinement. Future initiatives could focus on the adaptation of models to various linguistic and cultural environments, acknowledging the diverse ways people express suicidal thoughts across different languages and cultures. Furthermore, a holistic approach that integrates multiple data modalities, such as images, audio, or behavioral data, alongside textual information could enhance the detection process. It's also crucial to set up a framework that allows for the continuous evaluation and optimization of models, given the ever-changing nature of online communication patterns and user behaviors.

\section{Ethical Considerations}\label{sec:Ethics}
For this research, we obtained ethics approval from the research ethics board at the University of Ottawa. Moreover, the UMD dataset was used with authorization from its creators, and we adhered to the terms of use and ethical standards  \footnote{\href{http://users.umiacs.umd.edu/~resnik/umd_reddit_suicidality_dataset.html}{The University of Maryland Reddit Suicidality Dataset}} provided by them.

The use of LLMs for suicide-related synthetic datasets raises several ethical considerations.
Firstly, synthetic datasets should be generated in a way that avoids perpetuating or amplifying biases present in the original data. It is important to carefully examine the underlying data and the algorithms used in generating synthetic datasets to ensure fairness and mitigate potential biases.

Secondly, the process of generating synthetic datasets should be transparent and well-documented. It is essential to provide clear information about the methods used, assumptions made, and limitations of the synthetic data. This enables others to assess and evaluate the validity and appropriateness of using synthetic datasets.

Thirdly, to use synthetic datasets in sensitive applications or decision-making processes, accountability and liability should be considered. Care should be taken to understand the potential impact and consequences of decisions or actions based on synthetic data and establish mechanisms for addressing any negative outcomes or biases that may arise.

\bibliographystyle{unsrt}
\bibliography{Synthetic_data}
\begin{IEEEbiography}[{\includegraphics[width=1in,height=1.25in,clip,keepaspectratio]{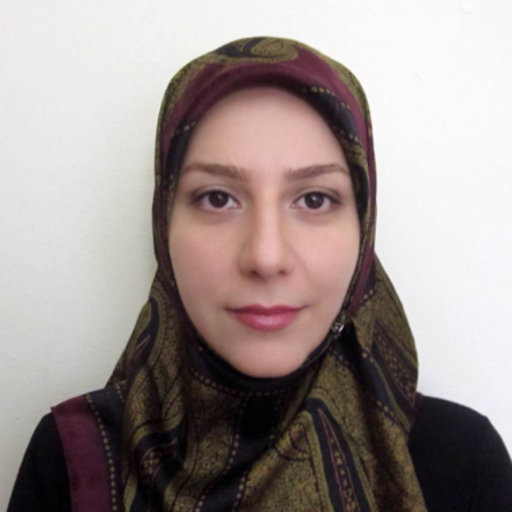}}]{Hamideh Ghanadian} is Ph.D. Candidate in Electrical Engineering and Computer Science at the University of Ottawa. She also completed her MASc degree in Electrical Engineering and Computer Science at the University of Ottawa in 2018. Her research focuses on Natural Language Processing,
Applied Machine Learning, Social Media Processing and explainability of AI systems. Her work particularly focuses on the application of natural language processing techniques on suicide and mental health detection on social media platforms and exploring the ways in which NLP can be used to better understand human psychology.
\end{IEEEbiography}
\begin{IEEEbiography}[{\includegraphics[width=1in,height=1.25in,clip,keepaspectratio]{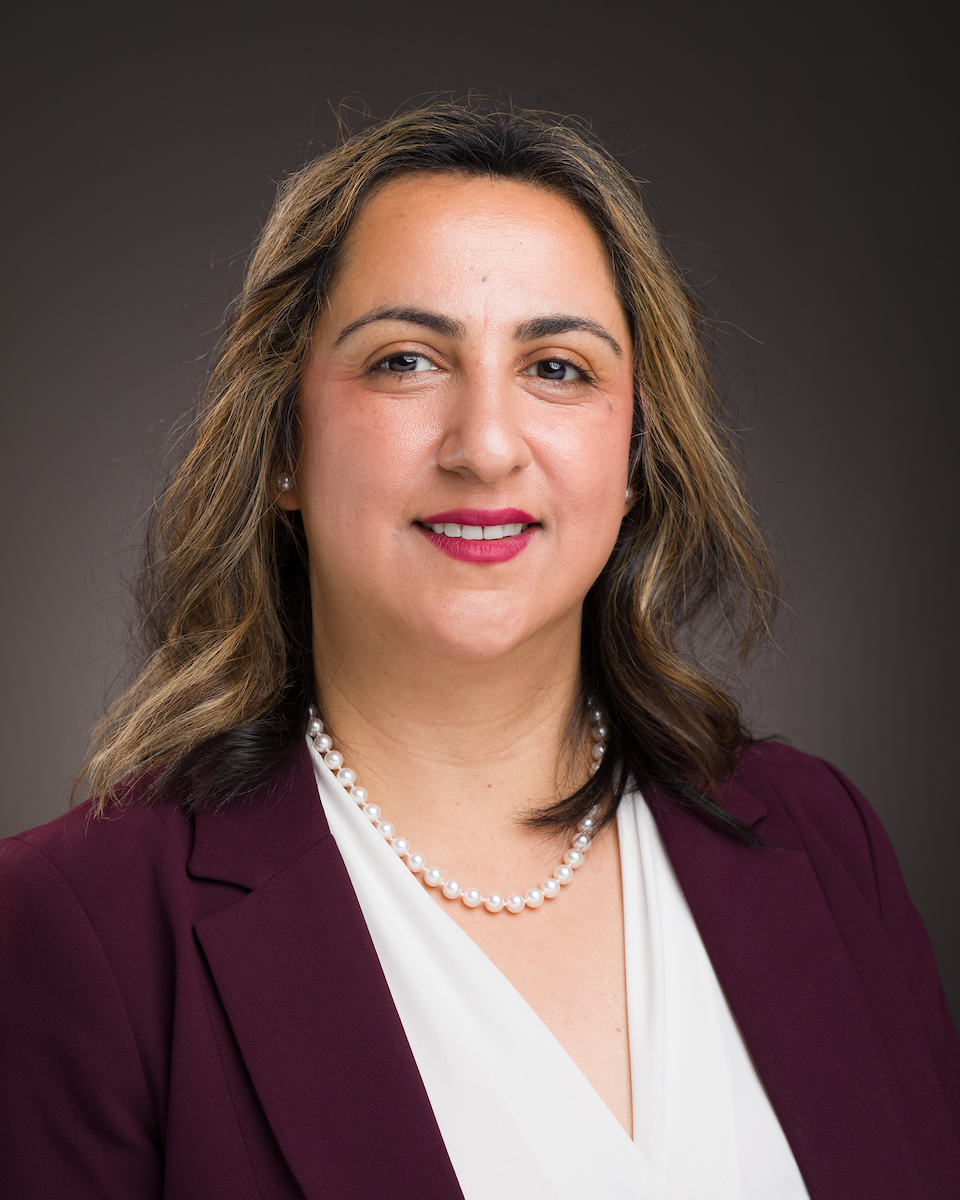}}]{Isar Nejadgholi} is a senior research scientist at the National Research Council Canada and an adjunct professor at the University of Ottawa. She completed her PhD in Artificial Intelligence at the AmirKabir University of Technology, Iran and her postdoctoral studies at the University of Ottawa, Canada, in 2016. Her research interests include machine learning applications, particularly natural language processing, social media data analysis and medical text processing. Her work also focuses on responsible AI, specifically on evaluating and improving the transparency and fairness of natural language processing systems. 
\end{IEEEbiography}
\begin{IEEEbiography}[{\includegraphics[width=1in,height=1.25in,clip,keepaspectratio]{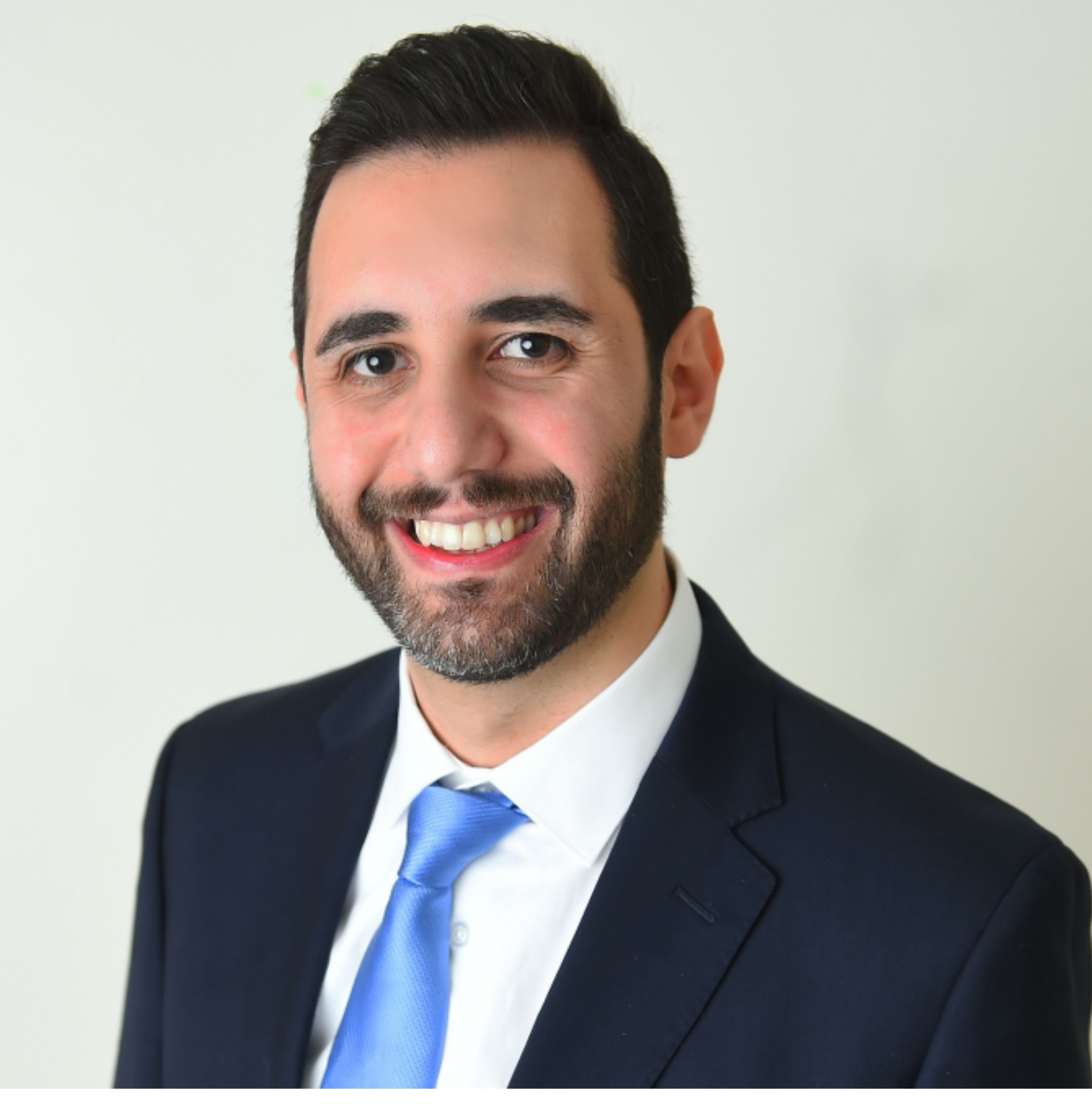}}]{Hussein Al Osman}
 is an Associate Professor at the School of Electrical Engineering and Computer Science at the University of Ottawa. He completed his Ph.D. in Electrical and Computer Engineering at the University of Ottawa in 2014. He leads the Multimedia Processing and Interaction group and is a member of the Multimedia Computing and the Distributed and Collaborative Virtual Environments Research laboratories. His research focuses on the application of artificial intelligence in affective computing and biomedical engineering. In particular, he is interested in the development of multi-modal affect recognition methods using deep artificial neural networks to estimate facial expressions and speech sentiment. He studies remote physiological signal measurement using video signals and applies this technology to biomedical and Human-Computer Interaction (HCI) applications. He conducts research in HCI, especially the development of serious games intended for physical rehabilitation and education. 
\end{IEEEbiography}

\EOD

\end{document}